\documentclass[]{article}
\usepackage{proceed2e}
\usepackage{natbib}
\usepackage{graphicx}
\usepackage{verbatim}
\usepackage{url}
\usepackage{subfigure}
\usepackage{algorithm, algorithmic}
\usepackage{amsmath, amsthm, amsfonts, amssymb}
\usepackage[usenames]{color}
\usepackage{tikz}
\usepackage{booktabs}
\usepackage{multirow}
\usepackage{rotating}
\usepackage{bm}
\usepackage{wrapfig}
\usepackage{multicol}
\usepackage{epsfig}

\newtheorem{theorem}{Theorem}

\title{Generalized Categorization Axioms}

\author{ {\bf Jian YU } \\
Beijing Key Lab
of Traffic Data Analysis and Mining\\
 Beijing Jiaotong University,Beijing, China\\
Email: jianyu@bjtu.edu.cn
}

\begin{document}

\maketitle

\begin{abstract}
Categorization axioms have been proposed to axiomatizing clustering results, which offers a hint of  bridging the difference between human recognition system
and machine learning through an intuitive observation: an object should be assigned to its most similar category.   However,  categorization axioms cannot be generalized
into a general machine learning system as  categorization axioms become trivial when the number of categories becomes one.
In order to generalize categorization axioms into general cases,  categorization input and categorization output are
reinterpreted by inner and outer category representation. According to the categorization reinterpretation,
two category representation axioms are presented. Category representation axioms  and categorization axioms can be combined into a generalized
categorization axiomatic framework, which accurately delimit the theoretical categorization constraints and overcome the shortcoming of categorization axioms. The proposed axiomatic framework not only
discuses categorization test issue but also reinterprets many results in machine learning in a unified way,  such as dimensionality reduction,
 density estimation, regression, clustering and classification.
\end{abstract}

{\bf Keywords:}  Similarity, Categorization, Category Representation, Dimensionality Reduction,Density Estimation, Regression, Clustering, Classification

\section{Introduction}

\par Up to now, many elegant but complex machine learning theories are developed for categorization, such as PAC theory \citep{valiant1984theory},
statistical learning theory \citep{vapnik2000nature} and so on. However, a six or seven year old child can easily and correctly categorize
many objects and does not understand about the above mentioned machine learning  theories. Therefore,
 there exists a clear gap between human recognition system and machine learning theories.  
 
 \par In \cite{JianYu2014Categorization},
 categorization axioms  have been proposed to axiomatizing clustering results, which  theoretically offers a hint of bridging
  the difference between human recognition system and machine learning by an intuitive observation:
  an object should be assigned to its most similar category.  Assumed that $c>1$ and the object representation of the input is the same as that of the output,
  \cite{JianYu2014Categorization} have proposed representation of clustering results and studied clustering results based on categorization axioms.
   However,  the proposed representation for clustering results in  \citep{JianYu2014Categorization} may be not available for many machine learning algorithms.
    For example,  when the number of categories becomes one, categorization axioms become trivial as they are always true. In the literature, many learning algorithms such as manifold
     learning and regression belong to one category learning problem. In order to generalize categorization axioms,
     categorization is needed to be further investigated.  
     
  \par According to the above analysis,
    several improvements on categorization axioms are made in this paper as follows:

1) A unified categorization representation is put forward and similarity operator and assignment operator are defined.

2) Category representation is axiomatized by two axioms, which includes existence axiom of category representation, and uniqueness axiom of category representation.

3) Three principles of developing categorization methods are investigated under new proposed categorization representation.

4) Categorization test is discussed by categorization test axiom and categorization robustness assumption.

5) Density estimation, regression, classification, clustering and dimensionality reduction  are  reinterpreted  by the proposed axioms.

The remainder of the paper is organized as follows: In section 2,  a unified categorization representation is discussed and
two axioms of category representation are presented. In section 3,
three categorization axioms are reinterpreted under new categorization representation.
In section 4, how to theoretically evaluate a categorization algorithm is discussed.
In section 5,  how to design a categorization method is discussed.
In section 6,  as applications of the proposed categorization axiomatic framework,
dimensionality reduction,density estimation, regression, clustering and classification are reinterpreted in a unified way.    The final section offers concluding remarks.

\section{Category Representation Axioms}\label{CRA}

In cognitive sciences,  a basic principle for human recognition system is  that an object should be assigned to its most similar category.
 For human being, membership explicitly represents  that an object is assigned to some category and must be observed by others,
 similarity between an object and a category  may be  implicit and may not be observed by others.
 In other words, human beings has two category representations for categorization,
  membership is explicit and is called outer category representation, similarity may be implicit and belongs to inner category representation.
   According to cognitive science, inner category representation for a category  is in the mind of human beings,
   which may be different from the outer category representation.
    Human being establish the relation between objects in the world and corresponding concepts in the mind by two category representations for categorization.
    For categories, a categorization algorithm should also have inner and outer category representations in order to reflect the relation
    between objects in the world and the corresponding categories as \cite{JianYu2014Categorization} have done for clustering results.
     Considered the limits of the proposed representation in   \citep{JianYu2014Categorization},
     we will reinterpret how to define  the  inner and outer category representation in a categorization algorithm in the following.

Any algorithm  has the input and the output.  For a categorization algorithm, the input is called categorization input and the output is called categorization result.
Categorization input  should have  inner and outer representation.  Inner categorization input is expected to be learned with respect to the outer categorization input.
 Similarly, Categorization output  should have  inner and outer representation.
  Inner categorization output is actually learned with respect to the outer categorization output.

The outer categorization  input is about the predefined categorization information of the sampling objects $O=\{o_1,o_2,\cdots,o_n \}$,  including the input object representation and the  corresponding outer category representation.

The input object representation  is  represented by $X=\{x_1,x_2,\cdots,x_n\}$   with $c$ subsets  $ X_1,X_2,\cdots,X_c$  , where  $x_k$ represents the $k^{th}$ object $o_k$, $X_i$ is a set that consists of all the objects of the $i^{th}$ category in the dataset $X$.  The outer category representation for the categorization input  can be represented by  $U =[u_{ik}]_{c \times n}$, $\forall i\forall k, u_{ik}\geq 0$ represents the membership of the object $x_k$ belonging to the $i^{th}$ category.
Hence, the outer categorization input   can be represented by  ($X,U$). More detailed can be seen in \citep{JianYu2014Categorization} .
When $U$ is known, one object should be assigned to the category with biggest membership. Therefore, assignment (outer referring) operator $\rightarrow$ can be defined as  $\vec{X}=\{\vec{x}_1,\vec{x}_2,\cdots,\vec{x}_n\}$, where $\vec{x}_k=\arg\max_i u_{ik}$.

Similarly, the  outer categorization result can be expressed by ($Y, V$), where $Y=\{y_1,y_2,\cdots,y_n\}$ represents the object representation for the output, $y_k$ also represents the $k^{th}$ object  $o_k$ , and $ Y_1,Y_2,\cdots,Y_c$  represents the corresponding input  $c$ subsets  $ X_1,X_2,\cdots,X_c$, $V$ is the outer category representation for the output, $V =[v_{ik}]_{c \times n}=[v_1, v_2, \cdots, v_n]$ is a partition matrix, $\forall i\forall k, v_{ik}\geq 0$ represents the membership of the object $y_k$ belonging to the $i^{th}$ category and $v_k=[v_{1k},v_{2k},\cdots,v_{ck}]^T$.  Similarly, assignment operator $\rightarrow$ is defined as $\vec{Y}=\{\vec{y}_1,\vec{y}_2,\cdots,\vec{y}_n\}$, where  $\vec{y}_k=\arg\max_i v_{ik}$. If $\vec{x}_k$,$\vec{y}_k$ are single value, $x_k$ belongs to the $\vec{x}_k^{th}$ category, $y_k$ belongs to the $\vec{y}_k^{th}$ category.  In common sense,  assignment operator $\rightarrow$ represents outer referring and reflects the external relation between the object and the category.

 As pointed out by \cite{JianYu2014Categorization},  the cognitive representation of a category is always supposed to exist, even in an implicit state when designing a categorization algorithm. For simplicity,  when the input $X=\{x_1,x_2,\cdots,x_n\}$ is categorized into $c$ subsets $X_1,X_2,\cdots, X_c$,  $\forall i$, $\underline{X_i}$ is supposed to be the cognitive representation of the $i^{th}$ category , and the  output $Y=\{y_1,y_2,\cdots,y_n\}$ is categorized into $c$ subsets $Y_1,Y_2,\cdots, Y_c$,  $\forall i$, $\underline{Y_i}$ is supposed to the cognitive representation of $i^{th}$ category.

As pointed out by \cite{JianYu2014Categorization}, when the cognitive representation for any category is defined, objects can be categorized based on the similarity between objects and categories. As the input is usually different from the output, the input category similarity mapping and the output category similarity mapping can be defined by computing the similarity between objects and  categories as follows.

 {\bf Input Category Similarity Mapping:} \\
  $Sim_X$: $X\times \{\underline{X_1},\underline{X_2},\cdots,\underline{X_c}\} \mapsto R_{+}$ is called category similarity mapping if an increase in $Sim_X(x_{k},\underline{X_{i}})$ indicates greater similarity between $x_{k}$ and $\underline{X_{i}}$, a decrease in $Sim_X(x_{k},\underline{X_{i}})$ indicates less similarity between $x_{k}$ and $\underline{X_{i}}$ .

 {\bf Output Category Similarity Mapping:} \\
  $Sim_Y$: $Y\times \{\underline{Y_1},\underline{Y_2},\cdots,\underline{Y_c}\} \mapsto R_{+}$ is called category similarity mapping if an increase in $Sim_Y(y_{k},\underline{Y_{i}})$ indicates greater similarity between $Y_{k}$ and $\underline{Y_{i}}$ , a decrease in $Sim_Y(y_{k},\underline{Y_{i}})$ indicates less similarity between $y_{k}$ and $\underline{Y_{i}}$ .

For input category similarity mapping, similarity (inner referring) operator $\sim$ can be defined as  $\widetilde{X}=\{\widetilde{x}_1,\widetilde{x}_2,\cdots,\widetilde{x}_n\}$,  where $\widetilde{x}_k=\arg\max_i Sim_X(x_{k},\underline{X_{i}})$. Similarly, for output category similarity mapping, similarity operator $\sim$ can be defined as  $\widetilde{Y}=\{\widetilde{y}_1,\widetilde{y}_2,\cdots,\widetilde{y}_n\}$,  where $\widetilde{y}_k=\arg\max_i Sim_Y(y_{k},\underline{Y_{i}})$. It is easy to know that if $\widetilde{y_k}$ is single value, the larger $Sim_Y(y_k, \underline{Y_{\widetilde{y}_k}})$, the better $Sim_Y$. Similarly, if $\widetilde{x}_k$ is single value, the larger $Sim_X(x_k, \underline{X_{\widetilde{x}_k}})$, the better $Sim_X$.

 Similarly, the input category dissimilarity mapping and the output category dissimilarity mapping can be defined as follows:

  {\bf Input Category Dissimilarity Mapping:} \\
  $Ds_X$: $X\times \{\underline{X_1},\underline{X_2},\cdots,\underline{X_c}\} \mapsto R_{+}$ is called  category dissimilarity mapping  if an increase in $Ds_X(x_{k},\underline{X_{i}})$ indicates less similarity between $x_{k}$ and $\underline{X_{i}}$, a decrease in $Ds_X(x_{k},\underline{X_{i}})$ indicates greater similarity between $x_{k}$ and $\underline{X_{i}}$.

 {\bf Output Category Dissimilarity Mapping:} \\
  $Ds_Y$: $Y\times \{\underline{Y_1},\underline{Y_2},\cdots,\underline{Y_c}\} \mapsto R_{+}$ is called  category dissimilarity mapping  if an increase in $Ds_Y(y_{k},\underline{Y_{i}})$ indicates less similarity between $y_{k}$ and $\underline{Y_{i}}$, a decrease in $Ds_Y(y_{k},\underline{Y_{i}})$ indicates greater similarity between $y_{k}$ and $\underline{Y_{i}}$.\footnote{In order to be consistent with the intuition, category similarity mapping and category dissimilarity mapping are usually supposed to be non negative in this section. In applications, category similarity mapping and category dissimilarity mapping can be negative.}

For input category dissimilarity mapping, similarity operator $\sim$ can be defined as  $\widetilde{X}=\{\widetilde{x}_1,\widetilde{x}_2,\cdots,\widetilde{x}_n\}$,  where $\widetilde{x}_k=\arg\min_i Ds_X(x_{k},\underline{X_{i}})$. Similarly, for output category dissimilarity mapping, similarity operator $\sim$ can be defined as  $\widetilde{Y}=\{\widetilde{y}_1,\widetilde{y}_2,\cdots,\widetilde{y}_n\}$,  where $\widetilde{y}_k=\arg\min_i Ds_Y(y_{k},\underline{Y_{i}})$.  If $\widetilde{x}_k$ is single value,  the less $Ds_Y(y_k, \underline{Y_{\widetilde{y}_k}})$, the better $Ds_Y$. Similarly, the less $Ds_X(x_k, \underline{Y_{\widetilde{x}_k}})$, the better $Ds_X$,where $\widetilde{x}_k$ is single value.
If  $\widetilde{x}_k^{th}$ and $\widetilde{y}_k$ are single value, $x_k$ is said to be similar to the $\widetilde{x}_k^{th}$ category, $y_k$ is said to be similar to the $\widetilde{y}_k^{th}$ category.  In daily life, similarity operator $\sim$ represents inner referring and established the latent relation between the object in the world and the cognitive category representation.

According to the above analysis, when the outer categorization input is ($X, U$), its corresponding inner categorization input can be represented by ($\underline{X},Sim_X$) or by ($\underline{X},Ds_X$), where $\underline{X}=\{\underline{X_1},\underline{X_2},\cdots,\underline{X_c}\}$.  For brevity,  ($X,U,\underline{X},Sim_X$) or by ($X,U,\underline{X},Ds_X$) is called the categorization input. ($\underline{X},Sim_X$) or  ($\underline{X},Ds_X$) are the inner category representation for the input, simply, called inner input.

Likely, when the outer categorization result is  ($Y, V$), its corresponding inner categorization result can be represented by ($\underline{Y},Sim_Y$) or by ($\underline{Y},Ds_Y$), where $\underline{Y}=\{\underline{Y_1},\underline{Y_2},\cdots,\underline{Y_c}\}$.  For brevity,  ($Y,V,\underline{Y},Sim_Y$) or by ($Y,V,\underline{Y},Ds_Y$) is called the categorization result.  ($\underline{Y},Sim_Y$) or  ($\underline{Y},Ds_Y$) are the inner category representation for the output, simply, called inner output. If  a categorization algorithm can explicitly output $\underline{Y}$, such a categorization algorithm can be called white  box.  If  a categorization algorithm can not explicitly output $\underline{Y}$ but only explicitly output $(Y,V)$, such a categorization algorithm can be called  black box.  If a categorization algorithm can explicitly output parts but not full of $\underline{Y}$, such a categorization algorithm can be called  grey box.

For a categorization algorithm, its outer input and outer output should have the corresponding inner category representations. Therefore, we call it Existence Axiom of Category Representation (ECR). More accurately, it can be expressed as follows:

{\bf 1)  ECR :}\\
For a categorization algorithm, if its outer input  is  ($X,U$) and its outer output is ($Y,V$), then there exists the corresponding inner input   ($\underline{X},Sim_X$) and  inner output   ($\underline{Y},Sim_Y$).

For a  categorization algorithm,  the input is  expected to have  the same category representation  as the output with respect to categorization.  $(\underline{X}, Sim_X)$  and the corresponding output $(\underline{Y},Sim_Y)$ is considered to have the same category representation with respect to categorization if $( \underline{X},\widetilde{X})=(\underline{Y},\widetilde{Y})$.  $(X,U)$ and $(Y,V)$ is considered to have the same category representation with respect to categorization if $\vec{X} =\vec{Y}$.
 Such an assumption is called Uniqueness Axiom of Category Representation (UCR), which can be expressed as follows:

{\bf 2)  UCR :}\\
 For a categorization algorithm,  its  categorization input  ($X,U,\underline{X},Sim_X$) and its corresponding categorization output   ($Y,V,\underline{Y},Sim_Y$) should satisfy $(\vec{X},\underline{X},\widetilde{X})=(\vec{Y},\underline{Y}, \widetilde{Y} )$.

ECR and UCR are called category representation axioms. ($X,U,\underline{X},Sim_X$) represents the category information by the outer information provider,  ($Y,V,\underline{Y},Sim_Y$) represents the category information by the categorization algorithm,  ($\underline{X},Sim_X$)  is  expected to be learned and represents the inner category representation of the outer information provider,  and ($\underline{Y},Sim_Y$) is actually learned and represents the inner category representation of the categorization algorithm. UCR offers the conditions that learning can be perfectly accomplished, which states that the categorization input and the categorization output have the same categorization semantics. Sometimes, $\vec{X}=\vec{Y}$ can be further enhanced
 into $U=V$. 

\section{Reinterpretation of Categorization Axioms}

According to \cite{JianYu2014Categorization},   categorization axioms includes Sample Separation Axiom (SS),  Category Separation  Axiom(CS) and Categorization Equivalency Axiom (CE).   For a categorization result ($Y,V,\underline{Y},Sim_Y$),  SS, CS and CE  can be reinterpreted by similarity operator and assignment operator as follows.

{\bf 1) SS:}  $\forall k \exists i (\tilde{y}_k=i)$

{\bf 2) CS:}   $\forall i\exists k(\tilde{y}_k=i))$

{\bf 3) CE:}   $\widetilde{Y}= \vec{Y}$

Moreover, we can prove Theorem \ref{propertieSS}.

\begin{theorem}
\label{propertieSS}
 If $\forall k \forall i$ $\forall j( (j\neq i) \rightarrow (Sim_Y(y_k,\underline{Y_i}) \neq Sim_Y(y_k,\underline{Y_j})))$, then SS must hold.
\end{theorem}


When a categorization result is not proper, there are some objects theoretically belonging to two and more categories. In other words, some objects are in the borderline of some category. Based on this fact,  boundary set can be defined as follows.\\
{\bf Boundary set:} For a categorization result (Y,V,$\underline{Y},Sim_Y$), the boundary set for ($Y,\underline{Y},Sim_Y$) is defined as follows.

$BS_{(Y,\underline{Y},Sim_Y)}=\{ y_{k} \mid  card(\widetilde{y}_k)>1 \}$

where $card(\widetilde{y}_k)$ represents the cardinality of a set $\widetilde{y}_k$.

Transparently,  the above analysis also holds for the  categorization input $(X,U,\underline{X},Sim_X)$.
Therefore,  $(X,U,\underline{X},Sim_X)$ should also satisfy SS, CS and CE.  For brevity, we will not repeat the similar result.  More interestingly, some relation can be established between UCR and CE by Theorem \ref{propertieCE}.
\begin{theorem}
\label{propertieCE}
 If the categorization input $(X,U,\underline{X},Sim_X)$ and the categorization result  (Y,V,$\underline{Y},Sim_Y$) satisfy CE,  then  $\widetilde{X}=\widetilde{Y}$ is equivalent to $\vec{X}=\vec{Y}$.
\end{theorem}

 As noted above, the input $x_k$ and the corresponding $y_k$  represent the same object $o_k$. Generally speaking, the input $x$ and the corresponding output $y$  represents the same object $o$,  therefore, it is naturally assume that there exists a mapping $\theta$ from $x$ to $y$,  i.e. $y=\theta(x)$.  When $\underline{X}=\underline{Y}$, it is easy to know that $Sim_Y(y_k,\underline{Y_i})=Sim_Y(\theta(x_k),\underline{Y_i})=Sim_Y(\theta(x_k),\underline{X_i})$. Hence, $Sim_X(x_k,\underline{X_i})$ can be defined by $Sim_Y(\theta(x_k),\underline{X_i})$. Therefore, it is easy to know that $\underline{X}=\underline{Y}$ implies that $\tilde{X}=\tilde{Y}$  when $Sim_X(x_k,\underline{X_i})$ is defined by $Sim_Y(\theta(x_k),\underline{X_i})$.  By Theorem \ref{propertieCE} and the above analysis, $\underline{X}=\underline{Y}$ play an essential role in UCR. In particular, when $c$=1, it is easy to know that  $\widetilde{X}=\widetilde{Y}$ and $\vec{X}=\vec{Y}$ hold trivially, $\underline{X}=\underline{Y}$ is the only meaningful requirement in UCR. Moreover, categorization axioms and UCR offer the conditions that category similarity mapping should satisfy, and states that the input category similarity mapping should be equivalent to the output category mapping with respect to categorization, which is called similarity assumption.  For categorization, it is very challenging to design a proper output category similarity mapping satisfying UCR and categorization axioms.  Usually,  the input category similarity mapping is not equivalent to the output category mapping with respect to categorization in practice, which is called similarity paradox. If similarity paradox occurs, the categorization error will be not zero. According to the above analysis, the key to solve similarity paradox is to keep $\underline{X}=\underline{Y}$ to be true. As a matter of fact, it is often true that $\underline{X}\neq \underline{Y}$. Therefore, how to solve similarity paradox is an eternal problem in categorization.

\begin{figure*}
  \centering
  \includegraphics[width=0.75\textwidth]{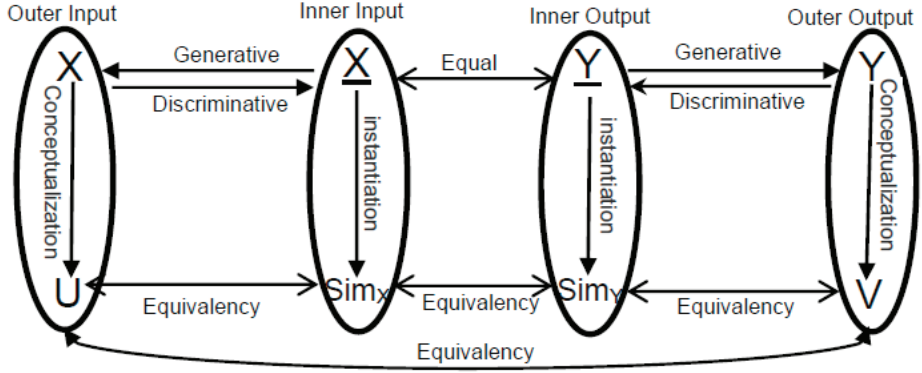}
  \caption{Relationship between a categorization input $(X,U,\underline{X},Sim_X)$ and its corresponding categorization result $(Y,V,\underline{Y},Sim_Y)$}\label{CSM}
\end{figure*}

 In summary, category representation axioms and categorization axioms have established the relationships among all the parts related to categorization input and categorization output, as shown in Figure \ref{CSM}.  UCR establishes the categorization equivalence between the input and the corresponding output.  Categorization axioms only establish the relationships between the outer representation and the corresponding inner representation and do not reflect the relation between the input and the output. If the object representation can be theoretically generated by  the corresponding  cognitive representation, then the corresponding  cognitive representation is called generative.  If the object representation can not be theoretically generated by  the corresponding  cognitive representation but can decide the corresponding  cognitive representation, then the corresponding cognitive representation is called  discriminative. If the  cognitive representation is generative,  the corresponding learning model is called generative model.  If the cognitive representation is generative,  the corresponding learning model is called discriminative model.

In particular, Let $X=Y$ and UCR be true,  $(\underline{X},Sim_X)$ and $(\underline{Y},Sim_Y)$ are exchangeable with respect to categorization.  Under such assumptions, $(X,U,\underline{X},Sim_X)$ can be used to represent the categorization results, where $(\underline{X},Sim_X)$ actually denotes $(\underline{Y},Sim_Y)$.  In \cite{JianYu2014Categorization}, ECR and UCR are implicitly assumed to be true, such an assumption makes CE  not be true as it is very difficult for $(\underline{Y},Sim_Y)$  to have the same categorization capacity as $(X,U)$ in practice, especially for $U$ is given a priori.

\section{Categorization Test}
 All the above analysis does not discuss how to evaluate the categorization result $(Y,V,\underline{Y},Sim_Y)$. Frankly speaking, it is very challenging to test the performance of a categorization algorithm.  When estimating the categorization performance, a test set $(X_T, U_T)$ is usually provided and $(X,U)$ is called the training set. According to the analysis in section \ref{CRA}, $(\underline{X_T}, Sim_{X_T})$ exists. Similarly, if $(X_T, U_T, \underline{X_T}, Sim_{X_T})$ is used the categorization input, the corresponding categorization output can be represented by $(Y_T, V_T, \underline{Y_T}, Sim_{Y_T})$.

 It is easy to know the test set and the training set are supposed to represent the same categorization for the same categorization algorithm. Therefore, Categorization Test Axiom can be expressed as follows:

 {\bf Categorization Test Axiom:} For a categorization algorithm, if its training test is $(X, U)$ and its test set is $(X_T, U_T)$, then $(\underline{X}, Sim_X)$=$(\underline{X_T}, Sim_{X_T})$.

 Certainly, categorization test axiom offers the prerequisite condition that a categorization algorithm has generalization ability, which is a demanding requirement for categorization.  It is easy to prove that categorization test axiom can infer the objects in the training set and the test set should be independent and identically distributed if objects are random variables.

 Usually, $\underline{X}$ only approximates $\underline{X_T}$.  Sometimes, the difference between $\underline{X}$ and $\underline{X_T}$ is so big that $\underline{X}$ and $\underline{X_T}$ cannot be considered to represent the same categorization.  In this case, the test result will be not credible and it can not be checked whether the corresponding categorization algorithm has generalization ability or not.

 In fact,$\underline{X}$ and $\underline{X_T}$ are unobservable and unknown, it is very difficult to measure the difference between $\underline{X}$ and $\underline{X_T}$.  Instead of measuring the difference between $\underline{X}$ and $\underline{X_T}$, one estimation method is to compute the difference between $(X, U)$ and $(X_T, U_T)$, the other estimation method is to compute the difference between $\underline{Y}$ and $\underline{Y_T}$ assuming that UCR holds or approximately holds at least.  Theoretically, the difference between $\underline{X}$ and $\underline{X_T}$ should be proportional to the difference between $\underline{Y}$ and $\underline{Y_T}$ in the ideal case.  Therefore, the categorization robustness assumption can be described as follows:

 {\bf Categorization Robustness Assumption:}  A categorization algorithm is called robust if there exist two constants $k_1$ and $k_2$ such that $k_1|\underline{Y}-\underline{Y_T}|\leq |\underline{X}-\underline{X_T}|\leq k_2|\underline{Y}-\underline{Y_T}|$, where $0<k_1\leq k_2$.

 Categorization robustness assumption demonstrates the global condition that the corresponding categorization algorithm has generalization ability when categorization test axiom does not hold. If categorization test axiom holds, a good categorization axiom should make $|\underline{Y}-\underline{Y_T}|$ as small as possible, When categorization test axiom dose not hold, it is very challenging to check whether or not categorization robustness assumption holds as $\underline{X}$ and $\underline{X_T}$ are usually not known.
 Therefore,  a substitutional method is to compute the distance between the outer representations. Such an idea leads to local categorization robustness assumption as follows:

{\bf Local Categorization Robustness Assumption:} A categorization algorithm is called locally robust if there exist two constants $k_1$ and $k_2$ such that
$k_1|(Y, V)-(Y_T, V_T)|\leq |(X, U)-(X_T, U_T)|\leq k_2|(Y, V)-(Y_T, V_T)|$, where $0<k_1\leq k_2$, $(X, U)$ is a training test and $(X_T, U_T)$ is a test set.

Transparently, if local categorization robustness assumption is satisfied with respect to $|(X, U)-(X_T, U_T)|<\varepsilon$ where $\varepsilon$ is a very small positive number, the corresponding algorithm can be stably evaluated in theory.  

\section{Design Principles of Categorization Methods}

When categorization axioms are proposed by \cite{JianYu2014Categorization}, three design principles
of clustering methods have also  been proposed by \cite{JianYu2014Categorization}. However,
three design principles of clustering methods proposed by \cite{JianYu2014Categorization}  need to be reinterpreted  when categorization is investigated.
It is easy to guess that five axioms are also useful for developing categorization methods when five axioms are proposed to deal with categorization algorithms.
Clearly, five axioms do not have equal importance when designing a categorization method.
 ECR only tells us how to represent the categorization input and the categorization output.
 CE is always supposed to be true for a categorization algorithm since the outer referring and the corresponding inner referring should represent the same referring,
  in a word, the explicit function of a categorization algorithm should be the same as its internally implemented function.
   As pointed by \cite{JianYu2014Categorization},  SS and CS offer a very low bar for clustering results.  Similarly, SS and CS are also loose requirements for categorization.
   UCR is far demanding as it requires three equivalence conditions are true simultaneously.
   Therefore,   three design principles of categorization methods can be inferred from SS, CS and UCR.
   In the following, we will carefully investigate such three principles respectively under the proposed axiomatic framework.

\subsection{Category Compactness Principle}

Theorem \ref{propertieSS} shows that the conditions of SS are nearly no requirement
 as the conditions of Theorem \ref{propertieSS} are often true in general case for a well designed category similarity.
 Following the same analysis in  \cite{JianYu2014Categorization}, SS should be enhanced into category compactness principle as follows:

{\bf Category Compactness Principle:}  A categorization method should make its categorization result as compact as possible.

Category compactness principle says that every category should be as much compact as possible.
Under the proposed representation of the categorization result, category compactness criterion can be defined as follows.

{\bf Category Compactness Criterion:} $J_{C}: \{Y,V\}\times \{\underline{Y},Ds_Y\} $  $\mapsto R_{+}$ is called category compactness criterion if the optimum of $J_{C}(Y,V,\underline{Y},Ds_Y)$ corresponds to the categorization result with the largest category compactness.

According to categorization axioms, category compactness criterion can be equivalently defined by $J_{C}(X,U,\underline{X},Ds_X)$. In the literature, it is often seen that  $J_{C}(X,U,\underline{X},Ds_X)=\sum_i\sum_k u_{ik}Ds_X(x_k,\underline{X_i})$.

As the relevance among $(X,U,\underline{X},Ds_X)$,  $J_{C}(X,U,\underline{X},Ds_X)$ can be further simplified into  $J_{C}(X,\underline{X},Ds_X)$ or $J_{C}(U)$.
Noticing the definition of category similarity mapping, category compactness principle is still available for categorization when $c=1$.

\subsection{Category Separation Principle}

If a categorization result $(Y,V,\underline{Y},Sim_Y)$ satisfies CS,
then $\forall 1\leq i\neq j\leq c,\underline{Y_i} \neq \underline{Y_j}$.
According to the same reason in \cite{JianYu2014Categorization}, CS can be enhanced into category separation  principle as follows:

{\bf Category Separation Principle:} A good categorization result should have the maximum distance between categories.

Under the proposed representation of the categorization  result, category compactness criterion can be defined as follows.

{\bf Category Separation Criterion:} \\
$J_{S}: \{Y,V\}\times \{\underline{Y_1},\underline{Y_2},\cdots,\underline{Y_c}\}  \mapsto R_{+}$ is called category separation criterion if the optimum of $J_{S}(Y,V, \{\underline{Y_1},\underline{Y_2},\cdots,\underline{Y_c}\})$ corresponds to the categorization result with maximal category separation.

Category separation principle requires that $c>1$.  In other words, when $c=1$, category separation principle is unavailable.
\subsection{Categorization Consistency Principle}

If the categorization input $(X,U,\underline{X},Sim_X)$  and its corresponding categorization result $(Y,U,\underline{Y},Sim_Y)$ satisfy UCR,
the categorization error is zero. However, even for human recognition systems, UCR can not be always guaranteed to be true.
Generally,  human recognition systems always try to make categorization error as small as possible.
Therefore, UCR is the most demanding requirement for categorization.
If UCR does not hold,  a reasonable categorization criterion should make UCR  hold as approximately as possible,
 which result in categorization consistency principle as follows:

{\bf Categorization Consistency Principle:} When UCR does not hold, a good categorization result should make UCR as approximately  correct as possible.

When UCR does not hold, categorization consistency principle can be used to design some categorization criterion as follows:

{\bf Categorization Consistency Criterion:} $J_{E}: \{X,\vec{X}, \underline{X}, \widetilde{X} \} \}\times \{ Y,\vec{Y}, \underline{Y}, \widetilde{Y} \} $ $\mapsto R_{+}$ is called categorization consistency criterion if the optimum of $J_{E}(X,\vec{X},\underline{X}, \widetilde{X}, Y,\vec{Y},\underline{Y}, \widetilde{Y})$ corresponds to the categorization result with the minimum difference between ($\vec{X},\underline{X}, \widetilde{X}$) and ($\vec{Y},\underline{Y}, \widetilde{Y} $).

Clearly, if UCR can not be true, categorization consistency principle should be the first principle
when designing a categorization algorithm no matter what the number of categories is.
Frankly speaking, it is not usually expected that ($\underline{X},Sim_X$)  and  ($\underline{Y},Sim_Y$) are  obtained simultaneously.
Usually, ($\underline{X},Sim_X$) is interchanged or approximated by ($\underline{Y},Sim_Y$) when designing a categorization algorithm.
In many categorization algorithms, UCR is supposed to be true but is not actually true.
Under such an assumption, category compactness principle and category separation principle should be used to design categorization methods.

\subsection{Occam's razor}
\label{occamsrazor}

For a specific categorization problem, there exists many categorization models. Category compactness principle,
category separation principle and categorization consistency principle just select the optimal parameters
in the candidate models with the same inner category representation, and cannot choose the optimal models among different inner category representations.
How to select an appropriate categorization model among different inner category representations?
 Occam¡¯s razor principle is a popular tool for human being to choose models among different representations,
 which states that "plurality should not be posited without necessity". Therefore, a simpler categorization model
 should be selected among the candidate models with the same performance.

What is a simple categorization model?
As the categorization problem can be represented by the categorization input $(X,U,\underline{X},Sim_X)$ and the corresponding
categorization output $(Y,V,\underline{Y},Sim_Y)$, a model with the simple categorization input and output will be considered simple.
 When $c$=1, then $\forall k, \vec{x}_k=1$ and $\forall k, \widetilde{x}_k=1$.
 Therefore, it is enough to study $\underline{X}$ and $\underline{Y}$ in order to obey UCR or its approximated version: categorization consistency principle,
  $(U,Sim_X,V,Sim_Y)$ can be omitted when designing a categorization model. If such an assumption holds, it can be considered as a simple categorization problem.
   Otherwise, if $c\geq 2$, assume $Y=X$, $V$ can be replaced by $Sim_Y$ because CE always hold, hence, $(Y,V,\underline{Y},Sim_Y)$ can
   be represented by $(\underline{Y},Sim_Y)$. Similarly, $(X,U,\underline{X},Sim_X)$ can be represented by $(X,U)$.
   In this case, it is enough to deal with $(X,U,\underline{Y},Sim_Y)$ for such a categorization problem.
    Clearly, it is also a simple categorization case.  Of course, such simplified categorization models can be further simplified by selecting simpler $\underline{Y}$.
     In summary, Occam's razor can be used to discuss categorization model complexity.
      In the following, we will study categorization models according to model complexity in the Occam's razor point of view.

\section{Applications}
In this section, we will study categorization models according to analysis in section \ref{occamsrazor}. When $c=1$, categorization becomes one category problem, including density estimation, regression  and some dimensionality reduction methods.  When $c>1$, categorization is multiple category problem, including clustering and classification. When $U$ is not know for $c>1$  before categorization, categorization is a clustering problem. when $U$ is known for $c>1$  before categorization, categorization is a classification problem. In the following, the above issues will be discussed based on the proposed axioms and principles.

\subsection{Unsupervised Dimensionality Reduction}

In the following, we will give several examples to show how to interpret dimensionality reduction methods based on the proposed axioms and principles.

For simplicity, assume that  $X=[x_{kr}]_{n\times p}$ are sampled from some underlying structure  in a  space with dimensionality $p$,  and such a sample can also be represented by $Y= [y_{kr}]_{n\times d}$ in a low dimensional space with dimensionality $d$, where $p>>d$.  Such a categorization problem is called dimensionality reduction.


If $U$ is not known, such a problem is called unsupervised dimensionality reduction. It is easy to know that unsupervised dimensionality reduction has the categorization input $(X,U, \underline{X}, Ds_X)$ and the categorization output $ (Y,V, \underline{Y}, Ds_Y)$.   Therefore, unsupervised dimensionality reduction can be considered a categorization problem. In this section, we further assume that $c=1$.  Under this assumption, it is easy to know that $\tilde X=\tilde Y$ and $\vec{X}= \vec{Y}$.  UCR only requires that $\underline{X}=\underline{Y}$. If UCR does not hold, categorization consistent principle naturally requires that  $\underline{X}$  approximates  $\underline{Y}$ as much as possible.  If UCR does hold,  category compactness principle implies that  the best $\underline{X}$ should make the underlying category the most compact.

{\bf  PCA\citep{pearson1901,hotelling1933analysis,abdi2010principal}:} Let $\underline{X}=\underline{Y}=\left [ \begin{array} {c}  x_0 \\  w_1 \\w_2\\ \cdots \\  w_d \\ \end{array} \right ]$ represent the ordered orthonormal basis $\{w_1,w_2,\cdots,w_d\}$  with the origin $x_0$, $Y=[y_{kr}]_{n\times d}$ are the coordinates of the objects $O=\{o_1,o_2,\cdots,o_n\}$ in the ordered orthonormal basis $\{w_1,w_2,\cdots,w_d\}$ with the origin $x_0$. Then we know that  $  w_iw_j^T=\delta_{ij}$, $\delta_{ij}=1$ if $i=j$, $\delta_{ij}=0$ if $i\neq j$, $y_{kr}=(x_k-x_0)w_r^T$, $x_0, w_i$ are  $1\times p$ vector.

 Let $Ds_X(x,\underline{X})=(x-x_0-\sum_i(x-x_0)w_i^Tw_i)(x-x_0-\sum_i(x-x_0)w_i^Tw_i)^T$ represent the dissimilarity between $x$ and the category representation $\underline{X}$ , it is easy to prove that $Ds_X(x,\underline{X})=(x-x_0)(x-x_0)^T-\sum_iw_i(x- x _0)^T(x-x_0)w_i^T$.  Obviously, if $x$ can be a linear combination of the ordered orthonormal basis $\{w_1,w_2,\cdots,w_d\}$  with the origin $x_0$, then  $Ds_X(x,\underline{X})=0$ means $x$ can be perfectly represented by $\underline{Y}$.  If $\forall x_k, Ds_X(x_k,\underline{X})=0$, then $\forall x_k$ have  the coordinates of the objects $O=\{o_1,o_2,\cdots,o_n\}$ in the ordered orthonormal basis $\{w_1,w_2,\cdots,w_d\}$ with the origin $x_0$ with zero residual.  In general cases, it is not true that $\forall x_k, Ds_X(x_k,\underline{X})=0$.

As UCR holds, category compactness principle will be used to seek the best $\underline{X}$, which means that a good $\underline{X}$ should minimize the objective function (\ref{PCA}) subject to  $\forall i\forall j, w_iw_j^T=\delta_{ij}$.
\begin{equation}\label{PCA}
\begin{split}
&\min_{\underline{X}}\sum_{k}Ds_X(x_k, \underline{X})\\
&=\sum_{k}(x_k-x_0)(x_k-x_0)^T\\
&-\sum_iw_i\sum_k(x_k-x_0)^T(x_k-x_0))w_i^T
\end{split}
\end{equation}

By Lagrange multiplier method, the objective function can be rewritten as (\ref{LagPCA})¡£
\begin{equation}\label{LagPCA}
\begin{split}
 &L=\sum_{k}(x_k-x_0)(x_k-x_0)^T\\
 &-\sum_iw_i\sum_k(x_k-x_0)^T(x_k-x_0)w_i^T\\
 &-\sum_i\lambda_i(w_iw_i^T-1)
 \end{split}
\end{equation}

The equations (\ref{derPCA}) can be obtained by differentiating (\ref{LagPCA}).

\begin{equation}\label{derPCA}
\begin{split}
&\frac{\partial L}{\partial x_0}=-2\sum_k(x_k-x_0)(I_p-\sum_iw_i^T w_i)=0\\
&\frac{\partial L}{\partial w_i}=2w_i\sum_k(x_k-x_0)^T(x_k-x_0)-2\lambda_iw_i=0
\end{split}
\end{equation}
Hence, the solution of minimizing (\ref{PCA}) subject to  $\forall i\forall j, w_iw_j^T=\delta_{ij}$ is as (\ref{SoluPCA}).
\begin{equation}\label{SoluPCA}
\begin{split}
&x_0=\sum_k\frac{x_k}{N}\\
&w_i\sum_k(x_k-x_0)^T(x_k-x_0)=\lambda_iw_i
\end{split}
\end{equation}

The equation (\ref{SoluPCA}) and minimizing (\ref{PCA}) can  introduce the traditional principle component analysis.
 The proposed axiomatic framework of categorization has offered a new interpretation of principle component analysis.

{\bf NMF\citep{Lee1999Learning}:}  Let $Y=H=[h_{kr}]_{n\times d}$, $\underline{X}=\underline{Y}=W=\left [ \begin{array} {c}  w_1 \\w_2\\ \cdots \\  w_d \\ \end{array} \right ]$ represent  the ordered basis $\{w_1,w_2,\cdots,w_d\}$, $Y=[h_{kr}]_{n\times d}$ are the coordinates of the objects $O=\{o_1,o_2,\cdots,o_n\}$ in the ordered  basis $\{w_1,w_2,\cdots,w_d\}$, where
 all the elements in $w_i$ are negative and $\forall k,r, h_{kr}$ are negative.

Let $Ds_X(x_k,\underline{X})=(x_k-\sum_ih_{ki}w_i)(x_k-\sum_ih_{ki}w_i)^T$. As UCR holds, category compactness principle will be used to seek the best $\underline{X}$, which means that a good $\underline{X}$ should minimize the objective function (\ref{NMF}).
\begin{equation}\label{NMF}
\begin{split}
&\min_{\underline{X}}\sum_{k}Ds_X(x_k, \underline{X})\\
&=\sum_{k}(x_k-\sum_ih_{ki}w_i)(x_k-\sum_i h_{ki} w_i)^T\\
&=\|X-HW\|^2
\end{split}
\end{equation}
Minimizing (\ref{NMF}) introduces nonnegative matrix factorization \citep{Lee1999Learning}.

{\bf CCA\citep{hotelling1936relations}:}
Let $\underline{X}=\frac{Xa^T}{|Xa^T|}$ and $\underline{Y}=\frac{Yb^T}{|Yb^T|}$, where $a$ is $1\times p$ vector, $b$ is $1\times d$ vector.
However, $\underline{X}=\underline{Y}$ does not hold in general, UCR is not true.  Therefore, we should use categorization consistence principle,  which means to  minimize the objective function (\ref{CCA1}).
\begin{equation}\label{CCA1}
\begin{split}
&\min_{a,b}L(\underline{X}, \underline{Y})=|\underline{X}- \underline{Y}|^2=\left |\frac{Xa^T}{|Xa^T|}- \frac{Yb^T}{|Yb^T|}\right |^2\\
&=2-2\frac{(Xa^T, Yb^T)}{|Xa^T||Yb^T|}
\end{split}
\end{equation}
Obviously, minimizing  (\ref{CCA1}) is equivalent to maximizing  (\ref{CCA2})
\begin{equation}\label{CCA2}
\frac{(Xa^T, Yb^T)}{|Xa^T||Yb^T|}=\frac{aX^TYb^T}{\sqrt{aX^TXa^T}\sqrt{bY^TYb^T}}
\end{equation}
Hence, canonical correlation analysis is introduced by maximizing (\ref{CCA2}).

 {\bf  LLE\citep{roweis2000nonlinear}:}  Let $\underline{X}=W_X=[w_{kl}]_{n\times n}$, $Ds_X(x_k,\underline{X} )$=$Ds_X(x_k, W)$=$|x_k-\sum_{j\in N(k)}w_{kj} x_j|^2$, where $\sum_l w_{kl}=1$,$w_{kl}\geq 0$, $w_{kl}= 0$ if $l \notin N(k)$,  $N(k)=\{j| x_j$ is  the neighbor of $x_k\}$.

As UCR holds, category compactness principle will be used to seek the best $\underline{X}$.  According to category compactness principle, a good category representation $\underline{X}=W$ should minimize the objective function (\ref{LLEW}):

\begin{equation}\label{LLEW}
   \min_W\sum_{k}Ds_X(x_k, W)=\sum_{k}|x_k-\sum_{j\in N(k)}w_{kj} x_j|^2
 \end{equation}

According to UCR, $\underline{X}=\underline{Y}$ implies that $\underline{Y}=W$. Set $Ds_Y(y_k,\underline{Y} )$=$Ds_Y(y_k, W)$=$|y_k-\sum_{j\in N(k)}w_{kj} y_j|^2$, category compactness principle  tells us that a good $Y$ should minimize the objective function (\ref{LLEY})
\begin{equation}\label{LLEY}
   \min_Y\sum_{k}Ds_Y(y_k, W)=\sum_{k}|y_k-\sum_{j\in N(k)}w_{kj} y_j|^2
 \end{equation}
 By this way, local linear embedding algorithm can be resulted by minimizing (\ref{LLEW}) and (\ref{LLEY}).

  {\bf MDS\citep{joseph1978multidimensional}:}  Let $\underline{X}=D_X=[d_{kl}^X]_{n\times n}$, $\underline{Y}=D_Y=[d_{kl}^Y]_{n\times n}$,where $d_{kl}^X=|x_k-x_l|$, $d_{kl}^Y=|y_k-y_l|$. It is easy to know that $\underline{X}=\underline{Y}$ cannot hold.  Therefore, categorization consistence principle will be used, which requires that a good $\underline{Y}$ should minimize the  objective function (\ref{MDS}).

\begin{equation}\label{MDS}
   \min_YL(\underline{X},\underline{Y})=L(D_X,D_Y)
 \end{equation}
Naturally, multidimensional scaling (MDS) algorithm can be introduced by minimizing the objective function (\ref{MDS}).

{\bf ISOMAP\citep{tenenbaum2000global}:}
Let $\underline{X}=D_X=[d_{kl}^X]_{n\times n}$, $\underline{Y}=D_Y=[d_{kl}^Y]_{n\times n}$,where $d_{kl}^X$ represents the geodesic distance between $x_k$ and $x_l$ , $d_{kl}^Y=|y_k-y_l|$.   It is impossible for $\underline{X}=\underline{Y}$.  Categorization consistence principle requires to minimize (\ref{MDS}).
 According to the above analysis,  multidimensional scaling (MDS) algorithm can be used to compute $Y$.

By this way, ISOMAP algorithm is introduced.\\

\subsection{Density Estimation}

If  $n$ points $x_1$, $ x_2$,$\cdots$, $x_n$  are sampled from a random variable with unknown probability density function $f$, then $f$  is expected to be constructed from the observed data $X=\{x_1, x_2,\cdots, x_n\}$,  which is called density estimation.   $f$ is called expected density function.

Set $X=Y$, $\underline{X}=f$, $\underline{Y}= \hat{f}$,  $U=[1,1,\cdots,1]_{1\times n}^T$,  $V=[1,1,\cdots,1]_{1\times n}^T$,   density estimation can be considered as  a categorization problem with the categorization input $(X,U, \underline{X}, Ds_X)$ and the categorization output $ (Y,V, \underline{Y}, Ds_Y)$, i.e. density estimation is a categorization problem with only one category. In the following, $\hat{f}$ is called density estimator. \\
Because all points belong to one category,  $\vec{U}=\vec{V}$ and $\widetilde{X}=\widetilde{Y}$ hold.  However,  $\underline{X}\neq \underline{Y}$. Therefore, UCR does not hold.\\
One method of density estimation is parametric estimation. If $p(x)$ is supposed to belong to the distribution family $p(x|\theta)$, density estimation will be transformed into estimating $\theta$. In other words, density estimation will become parametric estimation.
In this case, $\underline{X}=\theta$, $Ds_X(x,\theta)=-\log(p(x|\theta))$.  Let $\hat{\theta}$ be the estimation of $\theta$, we  have $\underline{Y}=\hat{\theta}$,$Ds_Y(x,\hat{\theta})=-\log(p(x|\hat{\theta}))$¡£
Therefore, category compactness principle requires to minimize intra category variance, which results in the objective function (\ref{densityestimation}).
\begin{equation}\label{densityestimation}
  \min_{\hat{\theta}}\sum^n_{k=1}Ds_Y(x_k,\hat{\theta})=\min_{\hat{\theta}}\sum^n_{k=1}-\log(p(x_k|\hat{\theta}))
\end{equation}
It is easy to know that maximum likelihood method is equivalent to minimizing (\ref{densityestimation}). \\
For example, let $\forall k,x_k\in R^p, x\in R^p$, $p(x|\hat{\theta})=\frac{1}{\sqrt{2\pi^p \hat{\sigma}^{2p}}}\exp[-\frac{1}{2}\frac{(x-\hat{\mu})^T(x-\hat{\mu})}{\hat{\sigma}^{2p}}]$, where
$\hat{\theta}=\{\hat{\mu}, \hat{\sigma}^{2p}\}$.
According to Equation (\ref{densityestimation}), the objective function (\ref{gaussidensityestimation}) can be inferred.
\begin{equation}\label{gaussidensityestimation}
\begin{split}
  &L=\sum^n_{k=1}-\log(p(x_k|\hat{\theta}))\\
  &=\sum^n_{k=1}(\frac{1}{2}\frac{|x_k-\hat{\mu}|^2}{\hat{\sigma}^{2p}}+\log\sqrt{2\pi^p\hat{\sigma}^{2p}})
\end{split}
\end{equation}
Minimizing (\ref{gaussidensityestimation}) can lead to the estimation of $\hat{\theta}=\{\hat{\mu}, \hat{\sigma}^{2p}\}$, where $\hat{\mu}=\sum^n_{k=1}\frac{x_k}{n}$, $\hat{\sigma}^{2p}=\sum^n_{k=1}\frac{|x-\hat{\mu}|^2}{n}$.\\

Another method of density estimation is non parametric estimation. In this method, less rigid assumptions are made about $f$. In the literature \citep{Silverman1986density},  non parametric density estimators include histograms, kernel density estimation, k-nearest neighbor method, etc.

Clearly, the key problem for density estimation is to estimate the difference between $\hat{f}$ and $f$. In theory, the minimum difference between $\hat{f}$ and $f$ should be expected according to categorization consistency principle. In the literature, theoretical conditions for $\hat{f}=f$ have been well studied in the limit point of view\citep{Silverman1986density}.

\subsection{Regression}
Generally, if $n$ points $(\hat x_1,f(\hat x_1))$, $(\hat x_2,f(\hat x_2))$,$\cdots$, $(\hat x_n,f(\hat x_n))$  are sampled from $(\hat x,f(\hat x))$ and $f$ is not known but is expected to be learned, such a problem is called regression.  Usually, $f$ is called expected regression function.

Set  $X=\left [ \begin{array} {cc} \hat x_1 & f(\hat x_1)\\ \hat x_2 & f(\hat x_2)\\ \cdots &\cdots\\ \hat x_n & f(\hat x_n)\\ \end{array} \right ]$,
$ Y=\left [ \begin{array} {cc} \hat x_1 & F(\hat x_1)\\ \hat x_2 & F(\hat x_2)\\ \cdots &\cdots\\ \hat x_n & F(\hat x_n)\\ \end{array} \right ]$, \\
$\underline{X}=(\hat x, f(\hat x))$, $\underline{Y}=(\hat x, F(\hat x))$, where $F$ is called predicted regression function, $U=[1,1,\cdots,1]_{1\times n}^T$,  $V=[1,1,\cdots,1]_{1\times n}^T$,  it is easy to know that regression has  the categorization input $(X,U, \underline{X}, Ds_X)$ and the categorization output $ (Y,V, \underline{Y}, Ds_Y)$. In other words, regression can be considered as a categorization problem with only one category. \\
Because all points belong to one category,  it is easy to prove that $\vec{U}=\vec{V}$ and $\widetilde{X}=\widetilde{Y}$.   However,  $\underline{X}\neq \underline{Y}$ in general cases.  Therefore, UCR does not hold. According to categorization consistency principle, a good  category representation $\underline{Y}$ should minimize the following objective function:
\begin{equation}\label{regression}
|\underline{X}-\underline{Y}|= D( f(\hat x), F(\hat x))
\end{equation}
It is impossible to directly compute $ D( f(\hat x), F(\hat x))$ as $f$ is unknown. Therefore, different definitions of $ D( f(\hat x), F(\hat x))$ lead to different regression algorithms.\\

For example, set $f(\hat x)\in R$ and $F(\hat x)=\hat w \hat x^T+b$. Assume that the dimensionality of $\hat x$ is $\tau$.  \\
If $D( f(\hat x), F(\hat x))
=\sum^n_{k=1} \|f(\hat x_k)-F(\hat x_k)\|^2$, linear regression is obtained by minimizing (\ref{regression}) if $n>>\tau$.\\
When  $n<<\tau$, it is easy to know that many feasible solutions can reach the same minimum of (\ref{regression}) as $n<<\tau$ implies that minimizing (\ref{regression}) faces singular problem.

How to select the optimal solution from many feasible solutions of minimizing (\ref{regression})? A natural idea is to select the feasible solution with minimum norm.

If using Euclidean norm, then $D( f(\hat x), F(\hat x))$ can be defined by $^n_{k=1} \|f(\hat x_k)-F(\hat x_k)\|^2+\lambda\|w\|^2$. Hence,  ridge regression is obtained by minimizing (\ref{regression}) .

When using $L_1$ norm, then  $D( f(\hat x), F(\hat x))$ can be defined by $\sum^n_{k=1} \|(f(\hat x_k)-F(\hat x_k)\|^2+\lambda\|w\|_{L_1}$. By this way,  Lasso regression is obtained by minimizing (\ref{regression}) \citep{Tibshirani1994Regression}.

\subsection{Clustering}

For clustering, $(X,U,\underline{X}, Sim_X)$ is called clustering input, $(Y,V,\underline{Y}, Sim_Y)$ is called clustering result.  Since $U$  and $V$ are unknown a priori  for clustering, it is always supposed that the inner input and the corresponding inner output should be the same.  It means that $(\underline{X},Sim_X)$=$(\underline{Y},Sim_Y)$. Under that assumption, it is assumed that $U=V$ for clustering. \\
When $Y=X$, the outer input and the outer output are the same, which implies that  $(X,U,\underline{X}, Sim_X)$ and $(Y,V,\underline{Y}, Sim_Y)$ are exchangeable with respect to clustering. In a word,  $(X,U,\underline{X}, Sim_X)$  also  represents  clustering result. As  $Sim_X$ and $Sim_Y$ are the same,  $Sim$ can denote $Sim_X$ and $Sim_Y$ for clustering.   Hence,  theoretical analysis  on clustering in \cite{JianYu2014Categorization} is also true under new categorization interpretation of this paper.

Even if $Y\neq X$, $(U,\underline{X}, \widetilde X)$=$(V,\underline{Y}, \widetilde Y)$ also holds for clustering, which means that ECR and UCR are still true.  In other words, ECR and UCR can always be omitted for clustering so that SS, CS and CE play more important role  for clustering. Frankly speaking, SS, CS and CE are enough for clustering.  Of course, when $Y\neq X$,  such clustering algorithms usually have feature extraction step  such as spectral clustering.

\subsection{Classification}

For classification, a category is called a class. In order to be consistent with the literature, $(X, U, \underline{X}, Sim_X)$ is called classification training input and categorization result $(Y,V, \underline{Y}, Sim_Y)$ is called classification training output in this section. More specifically, $(X, U)$ is called the training set, $(\underline{X}, Sim_X)$ is called the expected classifier, $(Y,V)$ is called the training result, $(\underline{Y}, Sim_Y)$ is called the learned classifier.
ECR and categorization axioms are usually true for classification. However, UCR is usually not true.

If UCR is true, the classification error will be zero. In practice, a classification method can only make its classification result to reach the minimum classification error, but usually its classification error is not zero. Therefore, UCR should be as a constraint for a classification problem.  In other words, when dealing with a classification problem, UCR should be true as much as possible in probability.

When $U$ is a proper partition, the corresponding classification problem is standard classification problem. When  $U$ is a overlapping partition, the corresponding classification problem is multi label classification problem.  For multi label classification, SS should be generalized as $\forall k \exists i ( i \in \widetilde{x}_k))$. Under such a generalization, multi label classification also follows SS.

When classification result $(Y,V, \underline{Y}, Sim_Y)$ is outputted, we can predict which category a new object should be assigned to.  In theory, the decision region for a classification result $(Y,V,\underline{Y},Sim_Y)$ can be defined as follows:

{\bf Decision Region:}\\ $\Omega =\{x | \exists i (\tilde y=i) \wedge (y=\theta(x)\}$.

In particular, the decision region for a class $\underline{Y_i}$ can be defined as follows:

{\bf Decision Region for a Class $\underline{Y_i}$:}\\ $\Omega_i =\{x | (\tilde y=i) \wedge (y=\theta(x)\}$.

Therefore, it is easy to know that $\cup_i \Omega_i=\Omega$.

The boundary for a classification result $(Y,V,\underline{Y},Sim_Y)$ can be defined as follows:

{\bf Boundary:} $\partial \Omega=\underline{\Omega}-\Omega^{\diamond}$, where $\underline{\Omega}$ represents the closure of $\Omega$, $\Omega^{\diamond}$ represents the interior of $\Omega$.

The training decision region can be defined as follows:

{\bf Training Decision Region:} $\Omega_{(\underline{Y},Sim_Y)} =\{x | \exists i \exists k((x\in \Omega_i) \wedge  (x_k\in \Omega_i) \wedge (  Sim_Y(\theta(x),\underline{Y_i})\geq Sim_Y(\theta(x_k),\underline{Y_i})) )\}$.

{\bf Training Decision Region for a class $\underline{Y_i}$:} $\Omega_{Y_i} =\{x | \exists k((x\in \Omega_i) \wedge  (x_k\in \Omega_i) \wedge (  Sim_Y(\theta(x),\underline{Y_i})\geq Sim_Y(\theta(x_k),\underline{Y_i})) )\}$.

The support vector for a classification result $(Y,V,\underline{Y},Sim_Y)$ can be defined as follows:

{\bf Support Vector:} If $x_k \in \partial \Omega_{(\underline{Y},Sim_Y)} $, then $x_k$ is called a support vector for the classification result $(Y,V,\underline{Y},Sim_Y)$.

The margin for a classification result $(Y,V,\underline{Y},Sim_Y)$ can be defined as follows:

 $\textbf{Margin}_{(\underline{Y},Sim_Y)}=\min_{i\neq j}d(\Omega_{Y_i},\Omega_{Y_j})$, where $ d(\Omega_{Y_i},\Omega_{Y_j})$  represents the distance between $\Omega_{X_i}$ and $\Omega_{Y_j}$.

 Transparently, decision region is used to judge which category one object should be assigned to, and the goal of the training decision region focuses on judging the quality of the classification result.

\subsubsection{Regression based Classification}

In the literature, one common idea of designing a classification algorithm is to transform classification to regression.
In order to do this, regression function needs  to be defined.  In the following, we will do this according to the proposed axiomatic framework.

Expected regression function can be defined as  $\rho(k)=\vec{x}_k$, where $U$ is a proper partition.  Under this circumstance, CE states that $\rho(k)=\widetilde{x}_k$ holds for a classification result. Similarly,  when $V$ is a proper partition, we set $H(k)=\vec{x}_k$, then CE guarantees that $H(k)=\tilde{y}_k$ holds.

Generally speaking, $x$ denotes the input object representation and $y$ denotes the corresponding output object representation. As $y=\theta(x)$,  $\rho(x)$ denotes  $\vec{x}$, the predicted regression function can be defined as  $h(x)=H(\theta(x))=H(y)=\tilde y$,  i.e $h(x)$ represents the predicted label. \\

 Set $X=\left [ \begin{array} {cc}  x_1 & \rho( x_1)\\x_2 & \rho( x_2)\\ \cdots &\cdots\\x_n & \rho( x_n)\\ \end{array} \right ]$, $ Y=\left [ \begin{array} {cc} x_1 & h( x_1)\\ x_2 & h( x_2)\\ \cdots &\cdots\\ x_n & h( x_n)\\ \end{array} \right ]$,
$\underline{X}=( x, \rho(x))$, $\underline{Y}=( x,h(x))$. Therefore, classification can be considered regression.

Using such denotation,  UCR requires that $\underline{X}=\underline{Y}$, which means $\forall x (\rho(x)=h(x))$. In practice, it is impossible  as $\rho(x)$ is not known a priori but only $ \rho(x_k)$ is known for $k\in \{1,2,\cdots,n\}$. Therefore, it is  natural to relax $\forall x (\rho(x)=h(x))$ as $P(\rho(x)\neq h(x))\leq \varepsilon$. PAC theory has provided a theoretical investigation on sufficient conditions of making $P(\rho(x)\neq h(x))\leq \varepsilon$ hold with a probability not less than $1-\delta$  ~\citep{valiant1984theory}.

  Therefore, UCR  is very important for classification. For developing a classification method, categorization consistency principle requires that $ \sum_{k=1}^nL(\rho(x_k), h(x_k))$ reaches the minimum, which is usually called minimizing empirical risk. Transparently, neural networks can be introduced by minimizing empirical risk. Usually, the more complexity of $h(x)$, the more small the empirical risk.  Therefore, the tradeoff between the empirical risk and the function complexity will lead to the structural risk \citep{vapnik2000nature}.\\
     In particular,  when c=2, $ \rho(x)\in \{1,2\}$.  Set $h(x)=1+\pi(x)$ and $L(\rho(x), h(x))=-(\rho(x)-1)\log(h(x)-1)-(2-\rho(x))log(2-h(x))=-(\rho(x)-1)\log(\pi(x))-(2-\rho(x))log(1-\pi(x))$ where $\pi(x)=\frac{exp(wx^T+b)}{1+exp(wx^T+b)}$, equation (\ref{regression})  tells us that the objective function of binomial logistic regression model \citep{hosmer2004applied} can be expressed as follows:
\begin{equation}\label{binomialregression}
\begin{split}
&\min_{\underline{Y}}\sum^n_{k=1}L(\rho(x_k),h(x_k))\\
&=-\sum^n_{k=1}(\rho(x_k)-1)(wx^T+b)\\
&+\sum^n_{k=1}\log(1+exp(wx^T+b)
\end{split}
\end{equation}

\subsubsection{Classification for $X=Y$}

However, many classification methods are not developed by transforming classification to regression.   In order to show this clearly, we simply assume $Y=X$, then classification result will  omit $Y$ as $X$ is known a priori. By analysis in Section \ref{occamsrazor}, it is enough to study $(X,U,\underline{Y},Sim_Y)$ under such simplification.Since $(X,U)$ is known for classification,  the simplest $\underline{Y}$ should be preferred according to Occam's razor. In the following, $U=[u_{ik}]_{c\times n}$ is a hard partition.\\

{\bf Example 1:} It is the simplest to set  $\underline{Y}=X$, which means that $\forall i,\underline{Y_i}=X_i$. Under such assumption, we do not know any essential information about $\underline{Y}$ except for $X$.   When $\forall i,\underline{Y_i}=X_i$, it is natural to set $Sim_Y(y,\underline{Y_i})=Sim_Y(x,\underline{Y_i})=\frac{|N_i(x)|}{K}$, $ N_i(x)= \{ x_l | x_l\in X_i \wedge x_l \in  \textrm{K-nearest neighborhood of  } x\}$.  Under the above assumption,  K-nearest neighbor classification method \citep{Coverhartnearestneighbor} is introduced.
It is easy to know that the categorization result of K-nearest neighbor classification follows categorization axioms in general cases.  Clearly, UCR does not hold for K-nearest neighbor classification in general.

{\bf Example 2:} Let $X=[x_{kr}]_{n\times p}$,  $Sim_Y(y,\underline{Y_i})=Sim_Y(x, \underline{Y_i})$ and  $g_i(x)=\log Sim_Y(x,\underline{Y_i})$ be discriminant function, SS requires that object $x$ is assigned to class $\underline{Y_i}$  if $g_i(x)=\max_j g_j(x)$. Occam's razor states that simpler $\underline{Y}$ is preferred. In theory,  if $\forall i,\underline{Y_i}$ is represented by $(w_i,w_{i0})$ where $w_i$ is a $1\times p$ vector and $w_{i0}\in R$,$g_i(x)=\log Sim_Y(x,\underline{Y_i})=w_ix^T+w_{i0}$. Such a categorization model is simpler, which is called linear discriminant analysis \citep{fisher1936use}. Transparently,  linear discriminant analysis also satisfies categorization axioms. 

{\bf Example 3:}   In particular, when c=2, it is natural to set $\forall i,\underline{Y_i}=(w_i,w_{i0})$. Occam's razor states that less parameters should be preferred. If set $\underline{Y_1}=(w,b-1)$ and $\underline{Y_2}=(-w,-b-1)$, the number of free parameters is the least. Therefore, $\underline{Y_1}=(w,b-1)$ and $\underline{Y_2}=(-w,-b-1)$ are the simplest linear classification representation according to Occam's razor. In this case, $g_1(x)=\log Sim_Y(x,\underline{Y_1})=wx^T+b-1$ and $g_2(x)=\log Sim_Y(x,\underline{Y_2})=-wx^T-b-1$. Set $wx^T+b-1\geq 0$ for $\forall x_k\in X_1$  and $-wx^T-b-1\geq 0$ for $\forall x_k\in X_2$,  categorization axioms hold. Therefore, category separation principle states that the optimal linear discrimination should keep the distance between the two parallel hyperplanes as large as possible when UCR holds, which leads to the famous support vector machine.

It is easy to know that the training decision region for support vector machine is $\Omega_{(\underline{Y},Sim_Y)} =\{x |wx^T+b-1\geq 0 ~\textrm{for}~ \forall x_k\in X_1  ~\textrm{and}~ -wx^T-b-1\geq 0 ~\textrm{for}~ \forall x_k\in X_2 \}$.   It is easy to prove that $\textbf{Margin}_{(\underline{Y},Sim_Y)}=\frac{2}{\sqrt{ww^T}}$. Larger $\textbf{Margin}_{(\underline{Y},Sim_Y)}$ means a better generalization for support vector machine, which has been proved by statistical learning theory \citep{vapnik2000nature}.

{\bf Example 4:}   Let $\underline{Y_i}=(w_i,w_{i0})$ where $1 \leq i\leq c-1$ but $\underline{Y_c}$ is unknown, and $ Sim_Y(x,\underline{Y_i})=\frac{exp(w_ix^T+w_{i0})}{1+\sum_{i=1}^{c-1}exp(w_ix^T+w_{i0})}$ if  $1 \leq i\leq c-1$, $ Sim_Y(x,\underline{Y_c})=\frac{1}{1+\sum_{i=1}^{c-1}exp(w_ix^T+w_{i0})}$. According to category compactness principle,
we should maximize the objective function  can be expressed as follows:
\begin{equation}\label{logisticregression}
\begin{split}
&\max_{\underline{Y_{1}},\underline{Y_{2}},\cdots,\underline{Y_{c-1}}}\sum^n_{k=1}\sum^c_{i=1}u_{ik}\log Sim_Y(x_k,\underline{Y_i})\\
&=\sum^n_{k=1}\sum^{c-1}_{i=1}u_{ik}(w_ix_k^T+w_{i0})\\
&-\sum^n_{k=1}\log(1+\sum_{i=1}^{c-1}exp(w_ix_k^T+w_{i0}))
\end{split}
\end{equation}

Such categorization model is called logistic regression\citep{CoX1958}. According to Occam's razor, logistic regression is more complex  than linear discriminant analysis. When $c>2$, logistic regression should not be considered as a regression model as no regression function can be defined.  Moreover, the $c^{th}$ class can be considered noise in logistic regression.

{\bf Example 5:} For a categorization model, we do not need a concrete form $\forall i, \underline{Y_i}$ explicitly. No matter how complicated $\underline{Y}$ is, it is enough to compute $Sim_Y$.   If $Sim_Y(y,\underline{Y_i})=Sim_Y(x, \underline{Y_i})=P(x,\underline{Y_i})$ and $v_{ik}=P(\underline{Y_i}|x_k)$, it is easy to know that Bayes classifier almost follows categorization axioms as the output $y=x \in Y_i$ just because $Sim_Y(x,\underline{Y_i})=\max_j Sim_Y(x,\underline{Y_j})=\max_j P(x,\underline{Y_j})=P(x,\underline{Y_i})$ and Bayes theorem guarantees that $\arg max_i P(x,\underline{Y_i})=\arg max_i P(\underline{Y_i}|x)$.   Therefore, it is very important for Bayes classifier  to estimate $Sim_Y$ or $V$ by $(X,U)$.

In particular, assume that $X=[x_{kr}]_{n\times p}$ represents $n$ objects and $x=[x_{*1},x_{*2},\cdots, x_{*p}]$ represents an object, where $x_{*r}$ is the $r^{th}$ feature.  According to categorization axioms, it is enough to calculate $\max_iP(x, \underline{Y_i})$ in order to classify $x$.  According to Occam's razor, we should select the simplest way to calculate $P(x, \underline{Y_i})$. The simplest way to estimate $P(x|\underline{Y_i})$ is to assume that each feature is conditionally  independent of every other features for given  $\underline{Y_i}$, then $P(x|\underline{Y_i})=\prod_{r=1}^{p}P(x_{*r}|\underline{Y_i})$. Let $P(\underline{Y_i})=\frac{card(X_i)}{n}$, then $Sim_Y(x,\underline{Y_i})$ can be computed by $P(\underline{Y_i})\prod_{r=1}^{p}P(x_{*r}|\underline{Y_i})$. Based on the above analysis, naive Bayes classifier \citep{duda1973pattern} can classify $x$ according to categorization axioms. Therefore, naive Bayes classifier is the simplest Bayes classifier with respect to Occam's razor. As $v_{ik}=P(\underline{Y_i}|x_k)$ can be computed and $V$ is a probability partition, Bayes classifier can be considered soft categorization.


{\bf Example 6:}   Let $Ds_Y(y,\underline{Y_i})=Ds_Y(x,\underline{Y_i})=R(\alpha_i| x)=\sum _{j=1}^c \lambda_{ij}P(\underline{Y_j}|x)$, where the action $\alpha_i$ denotes the decision to assign the output $y$ to class $\underline{Y_i}$ and $\lambda_{ij}$ denotes the cost incurred for taking the action $\alpha_i$ when the input $x$ belongs to $\underline{Y_j}$ . Transparently, the categorization result of minimum risk classification almost abides by categorization axioms.

{\bf Example 7:}   Let $Sim_Y(y,\underline{Y_i})=Sim_Y(x, \underline{Y_i})=U(\alpha_i| x)=\sum _{j=1}^c U_{ij}P(\underline{Y_j}|x)$, where the action $\alpha_i$ denotes the decision to assign the output $y$ to class $\underline{Y_i}$ and $U_{ij}$ measures how good it is to take the action $\alpha_i$ when the input $x$ belongs to $\underline{Y_j}$. Maximum expected utility classifier also almost follows categorization axioms.

{\bf Example 8:} In the above examples, $\forall i, \underline{Y_i}$ is represented by one unique prototype, no matter what implicit or explicit.  If assume that $\forall i, \underline{Y_i}$ can be represented by several prototypes, such a classifier is more complex. In decision tree classifier, $\forall i, \underline{Y_i}$ usually is represented by several mutual exclusive rules. It can be proved that decision tree classifier also follows categorization axioms.


\subsubsection{Classification for $X\neq Y$}
When $X\neq Y$ with $p>d$,  supervised dimensionality reduction is proposed to deal with the corresponding categorization. When  $X\neq Y$ with $p<d$, kernel methods are proposed for categorization. In the following, we will discuss them respectively.

{\bf Supervised Dimensionality Reduction}\\

For $X\neq Y$ with $p>d$, it is easy to know that $y=\theta(x)$ such that $\forall k, y_k=\theta(x_k)$.  The simplest $\theta$ is a projection mapping. If $\theta()$ is a projection mapping, supervised dimensionality reduction becomes feature selection.  Feature selection methods can be easily interpreted by categorization consistency principle.

If $\theta$ is not a projection mapping, the simplest $\theta$ is a linear mapping from $R^p$ to $R$. If there exists a direction $w$ such that all categories in $(X,U)$  can be linearly separable when  all points in $(X,U)$ is vertically projected into the direction $w$, we set   $\underline{Y_i}=\underline{X_i}=v_i w^Tw$, where $w$ is a $1\times p$ vector.   $Y=[z_{k}]_{n\times 1}$,where $ww^T=1$,$z_k=x_kw^T$, $v_i=\frac{\sum_{x_k\in X_i}x_k}{\mid X_i\mid}$.
$ Ds_X(x,\underline{X_i})$=$(xw^Tw-\underline{X_i})(xw^Tw-\underline{X_i})^T$=$w(x-v_i)^T(x-v_i)w^T$ ,
$Ds_Y(z,\underline{Y_i})$=$(zw-\underline{Y_i})(zw-\underline{Y_i})^T$ , it is easy to know that
$Ds_X(x,\underline{X_i})=Ds_Y(z,\underline{Y_i})$.

According to category compactness principle, we need to minimize $\sum_i\sum_{x_k\in X_i}Ds(x_k,\underline{X_i})=nwS_Ww^T$. According to category separation principle, we need to maximize $\sum_{i=1}^c|X_i|w(v_i-\overline{x})^T(v_i-\overline{x})w^T$=$nwS_Bw^T$,
where $\overline{x}=n^{-1}\sum_{k=1}^n x_k$.  Combining the above two functions, $\frac{wS_Ww^T}{wS_Bw^T}$ should be minimized, which leads to the generalized Fisher linear discriminant analysis.

 In particular, when $c=2$, it is easy to prove that $(\underline{X_1}-\underline{X_2})(\underline{X_1}-\underline{X_2})^T=w(v_1-v_2)^T(v_1-v_2)w^T=wS_Bw^T$.  Since $|X_1|w(v_1-\overline{x})^T(v_1-\overline{x})w^T+|X_2|w(v_2-\overline{x})^T(v_2-\overline{x})w^T
=\frac{|X_1||X_2|^2}{|X|^2}w(v_1-v_2)^T(v_1-v_2)w^T+\frac{|X_1|^2|X_2|}{|X|^2}w(v_1-v_2)^T(v_1-v_2)w^T
=\frac{|X_1||X_2|}{|X|}w(v_1-v_2)^T(v_1-v_2)w^T$,  it is easy to know that to minimize  $w(v_1-v_2)^T(v_1-v_2)w^T$ is equivalent to minimize $\sum_{i=1}^{2}|X_i|w(v_i-\overline{x})^T(v_i-\overline{x})w^T$,  Therefore, when $c=2$, generalized Fisher linear discriminant analysis becomes Fisher linear discriminant analysis.  Certainly,  Fisher linear discriminant analysis follows UCR if $(X,U)$ is linear separable in a direction $w$.

{\bf Kernel Methods}

For $X\neq Y$ with $p<d$,  assume that $Y$ is linearly separable and but $X$ is not linearly separable,  it is easy to know that $\theta()$ is a nonlinear mapping such that $\forall k,y_k=\theta(x_k)$.  Sometimes, the dimensionality of $Y$ is infinite. In this case, it is impossible to decide $\theta()$ by $(X,U)$ and $(Y,V)$.  Fortunately, when $(\underline{Y},Sim_Y)$ is obtained,  $(\underline{X},Sim_X)$ can be obtained by the kernel function $K(x,x_k)=(\theta(x),\theta(x_k))$, where $(\theta(x),\theta(x_k))$ represents the inner product.

By defining $K(x,x_k)$, most categorization algorithms can be reinvented in kernel methods. Interested readers can read the article \citep{ScholkopfBernhard2011Learning}.

In summary, classification models almost follow categorization axioms.  But different classification models have different model complexity. It should be pointed out that a complex model may be easily interpreted but a simple one may be difficult to be interpreted. Sometimes, a simple categorization model is very difficult to be discovered especially when it is not easy to be interpreted.

\section{Discussion and Conclusions}

\cite{JianYu2014Categorization} have presented categorization axioms based on the assumption that
 any category should have two kinds of representation.
 The main drawback of \citep{JianYu2014Categorization} is to ignore the clustering input by implicitly assuming
 the the clustering result and the clustering input should have the same category representation.
 However, the input and the output may not have the same category representation, even for some clustering algorithms.
 Therefore, categorization axioms cannot directly be applied to a general learning algorithm. In particular, categorization axioms assume that the number of categories is greater than one,
 which is invalid for regression and manifold learning.

 In order to generalize categorization axioms into general categorization methods,  we represent categorization problems
 by redefining categorization input as $(X,U,\underline{X},Sim_X)$ and categorization result as $(Y,V,\underline{Y},Sim_Y)$.
 Based on this proposed representations of categorization input and categorization result,
 similarity (inner referring) operator and assignment (outer referring) operator are defined.
 Such two proposed operators  are helpful not only for presenting UCR  but also for reinterpreting categorization axioms.
 ECR, UCR, SS,CS and CE indeed delimit the theoretical constraints for categorization.
 In particular, UCR  offers the theoretical constraints for a perfect categorization algorithm,
 which guarantees that expected to be learned is equivalent to actually learned, i.e. there are no gap between teaching and learning.
 More interestingly, if taking $(X,U,\underline{X},Sim_X)$ and $(Y,V,\underline{Y},Sim_Y)$ as a conversation between two persons,
 CE states that the outer category representation is equivalent to the inner category representation with respect to categorization,
 which  is consistent with maxim of quality in conversation: do not say what you believe to be false \citep{Grice1975}.
 UCR states that the input and the output should refer to the same categorization, which is also consistent with maxim of relation in conversation:
 make your contribution relevant \citep{Grice1975}. When a dialogue can be efficiently carried out, UCR and CE should be true in daily life.

\begin{figure}
  \centering
  \includegraphics[width=0.4\textwidth]{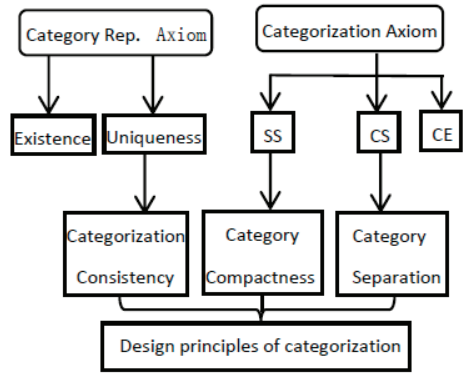}
  \caption{Relationship between Axioms and design principles for categorization}\label{Axiom}
\end{figure}

As the same as \cite{JianYu2014Categorization},
a clustering result satisfying SS and CS cannot be guaranteed to be a good clustering result as SS and CS are too weak.  Similarly,
when developing a categorization algorithm,  SS and CS also need to be enhanced, which respectively result in  the category compactness principle
and the category separation principle under new proposed representation.  In this paper, it is proposed that a categorization method should follow UCR in theory.
However, UCR is too demanding for a categorization method.  In many cases,  UCR cannot hold and needs to be relaxed,
which can lead to one design principle of categorization methods:
 the categorization consistency principle.
 The relation between the proposed axioms and design principles for categorization can be shown in Figure \ref{Axiom}.

After the learning process, how to evaluate the categorization algorithm is very important.
Categorization test axiom provides the prerequisite condition that the performance of the categorization algorithm can be evaluated and local
categorization robustness assumption has offered the condition that the performance of the categorization algorithm can be guaranteed to be stable.

When $c=1$, ECR, SS, CS and CE trivially hold, but UCR offers the theoretical condition for categorization.
When $c=1$, categorization becomes some dimensionality reduction methods, density estimation,
and regression. Some dimensionality reduction methods, density estimation and regression and can be introduced by UCR or
its approximated version (categorization consistency principle), such as principal component analysis,
nonnegative matrix factorization, canonical correlation analysis, local linear embedding, multidimensional scaling, Isomap,
parametric density estimation, nonparametric density estimation, linear regression, ridge regression and lasso, and so on.
Theoretically, when $c=1$, categorization mainly discusses how to represent a category, which lays on a foundation for categorization with $c>1$.\\
When $U$ is not known a priori for $c>1$, categorization becomes clustering.
ECR,UCR are always supposed to be true for any clustering algorithm in order to simplify clustering process.
Consequently, clustering result and clustering input are exchangeable when $X=Y$. Therefore, SS, CS and CE are enough for clustering when $X=Y$.
 Therefore, theoretical analysis of clustering  in \citep{JianYu2014Categorization}  still is true when $X=Y$.

When $U$ is known a priori for $c>1$, categorization becomes classification.
As for classification, ECR and CE are always true for a classification result, but SS and CS are true for a proper classification result and UCR just holds
for a classification result with zero error.  Therefore, SS, CS and UCR are more important constraints
for classification. With respect to a classification result $(Y,V,\underline{Y},Sim_Y)$, decision region,training decision region and margin are defined by SS.
 For categorization methods, category compactness principle can result in K-nearest classification,linear discriminant analysis,support vector machine,
 logistic regression, Bayesian classification, Minimum risk classification, Maximum expected utility classification,decision tree,etc.
  Category separation principle can lead to support vector machine and Fisher linear discriminant analysis.
   Categorization consistency principle can lead to empirical risk and structural risk, which can result in neural networks and binomial logistic regression model.

UCR, SS, CS and CE play different roles in different categorization algorithms but all have something to do with similarity.
 It is well known that that similarity plays a key role in human recognition system \citep{murphy2004big,UlrikeHahn2014Similarity}.
 Furthermore,  \cite{kloos2008s}  revealed that children represent categories based on similarity and  similarity-based category representation is a development default.
  The proposed axiomatic framework indeed establishes the bridge between cognitive science and machine learning through similarity (inner referring) operator.

More interestingly, the proposed categorization frame clearly shows the range that a categorization algorithm can be reasonably applied.
If the inner category representation is reasonable for the outer input, the corresponding categorization algorithm is feasible. Otherwise, more suitable inner category
representation should be used, which certainly introduces other categorization algorithm.  The analysis of categorization algorithms in this paper shows  that the design of cognitive category representation really needs powerful imagination as the cognitive category representations in the existing categorization algorithms are so diverse. In theory, a powerful categorization algorithm seems to have a powerful cognitive category representation.  

It should be pointed out that there are many open questions needed to be done in the proposed axiomatic framework in the future.
For example,  how to design an appropriate cognitive category representation for a specific categorization algorithm? When $c \geq 2$, how to solve similarity paradox?
What conditions can make categorization robustness assumption hold?
 When $(X,U)$ is partial known or noise, what is the relation between categorization axioms and categorization algorithms?


\subsubsection*{Acknowledgements}
Zongben Xu, Xinbo Gao, Wensheng Zhang, Baogang Hu, Jiangshe Zhang, Jufu Feng, Shaoping Ma, Qing He, Xuegang Hu, Liping Jing,  Bianfang Chai,  Jia Li  and all my colleagues in CAAI Machine Learning Technical Committee are appreciated very much,  their valuable discussions and suggestions have greatly improved the presentation of the this paper.  This work was supported by the NSFC grant (61370129), Ph.D Programs Foundation of Ministry of Education of China (20120009110006), PCSIRT(IRT 201206), Beijing Committee of Science and Technology,China(Grant No. Z131110002813118).

\bibliographystyle{apa}
\bibliography{bibfile}

\begin{thebibliography}{}

\bibitem[\protect\astroncite{Abdi and Williams}{2010}]{abdi2010principal}
Abdi, H. and Williams, L.~J. (2010).
\newblock Principal component analysis.
\newblock {\em Wiley Interdisciplinary Reviews: Computational Statistics},
  2(4):433--459.

\bibitem[\protect\astroncite{Cover and Hart}{1967}]{Coverhartnearestneighbor}
Cover, T. and Hart, P. (1967).
\newblock Nearest neighbor pattern recognition.
\newblock {\em IEEE Transactions on Information Theory}, 13(1):21--27.

\bibitem[\protect\astroncite{Cox}{1958}]{CoX1958}
Cox, D. (1958).
\newblock The regression analysis of binary sequences (with discussion).
\newblock {\em J. Roy. Stat. Soc. B}, 20:215--242.

\bibitem[\protect\astroncite{Duda et~al.}{1973}]{duda1973pattern}
Duda, R.~O., Hart, P.~E., et~al. (1973).
\newblock {\em Pattern classification and scene analysis}, volume~3.
\newblock Wiley New York.

\bibitem[\protect\astroncite{Fisher}{1936}]{fisher1936use}
Fisher, R.~A. (1936).
\newblock The use of multiple measurements in taxonomic problems.
\newblock {\em Annals of eugenics}, 7(2):179--188.

\bibitem[\protect\astroncite{Grice}{1975}]{Grice1975}
Grice, P. (1975).
\newblock Logic and conversation.
\newblock In {\em P.Cole and J. Morgan eds. Syntax and Semantics, vol.3, New
  York}. Academic Press.

\bibitem[\protect\astroncite{Hahn}{2014}]{UlrikeHahn2014Similarity}
Hahn, U. (2014).
\newblock Similarity.
\newblock {\em Wiley Interdisciplinary Reviews: Cognitive Science},
  5(3):271--280.

\bibitem[\protect\astroncite{Hosmer~Jr and Lemeshow}{2004}]{hosmer2004applied}
Hosmer~Jr, D.~W. and Lemeshow, S. (2004).
\newblock {\em Applied logistic regression}.
\newblock John Wiley \& Sons.

\bibitem[\protect\astroncite{Hotelling}{1933}]{hotelling1933analysis}
Hotelling, H. (1933).
\newblock Analysis of a complex of statistical variables into principal
  components.
\newblock {\em Journal of educational psychology}, 24(6):417.

\bibitem[\protect\astroncite{Hotelling}{1936}]{hotelling1936relations}
Hotelling, H. (1936).
\newblock Relations between two sets of variates.
\newblock {\em Biometrika}, pages 321--377.

\bibitem[\protect\astroncite{Kloos and Sloutsky}{2008}]{kloos2008s}
Kloos, H. and Sloutsky, V.~M. (2008).
\newblock What's behind different kinds of kinds: Effects of statistical
  density on learning and representation of categories.
\newblock {\em Journal of Experimental Psychology: General}, 137(1):52.

\bibitem[\protect\astroncite{Kruskal and
  Wish}{1978}]{joseph1978multidimensional}
Kruskal, J.~B. and Wish, M. (1978).
\newblock {\em Multidimensional scaling}, volume~11.
\newblock Sage.

\bibitem[\protect\astroncite{Lee and Seung}{1999}]{Lee1999Learning}
Lee, D. and Seung, H. (1999).
\newblock Learning the parts of objects by non negative matrix factorization.
\newblock {\em Nature}, 401(6755):788--791.

\bibitem[\protect\astroncite{Murphy}{2004}]{murphy2004big}
Murphy, G.~L. (2004).
\newblock {\em The big book of concepts}.
\newblock MIT press.

\bibitem[\protect\astroncite{Pearson}{1901}]{pearson1901}
Pearson, K. (1901).
\newblock On lines and planes of closest fit to systems of points in space.
\newblock {\em Philosophical Magazine}, 2(11):559�C572.

\bibitem[\protect\astroncite{Roweis and Saul}{2000}]{roweis2000nonlinear}
Roweis, S.~T. and Saul, L.~K. (2000).
\newblock Nonlinear dimensionality reduction by locally linear embedding.
\newblock {\em Science}, 290(5500):2323--2326.

\bibitem[\protect\astroncite{Scholkopf and
  Smola}{2011}]{ScholkopfBernhard2011Learning}
Scholkopf, B. and Smola, A.~J. (2011).
\newblock Learning with kernels: Support vector machines, regularization,
  optimization, and beyond.
\newblock {\em Journal of the American Statistical Association}, 98(3):781.

\bibitem[\protect\astroncite{Silverman}{1986}]{Silverman1986density}
Silverman, B. (1986).
\newblock {\em Density Estimation for Statistics and Data Analysis}.
\newblock Chapman \& Hall/CR,New York.

\bibitem[\protect\astroncite{Tenenbaum et~al.}{2000}]{tenenbaum2000global}
Tenenbaum, J.~B., De~Silva, V., and Langford, J.~C. (2000).
\newblock A global geometric framework for nonlinear dimensionality reduction.
\newblock {\em Science}, 290(5500):2319--2323.

\bibitem[\protect\astroncite{Tibshirani}{1994}]{Tibshirani1994Regression}
Tibshirani, R. (1994).
\newblock Regression shrinkage and selection via the lasso.
\newblock {\em Journal of the Royal Statistical Society, Series B},
  58(1):267--288.

\bibitem[\protect\astroncite{Valiant}{1984}]{valiant1984theory}
Valiant, L.~G. (1984).
\newblock A theory of the learnable.
\newblock {\em Communications of the ACM}, 27(11):1134--1142.

\bibitem[\protect\astroncite{Vapnik}{2000}]{vapnik2000nature}
Vapnik, V. (2000).
\newblock {\em The nature of statistical learning theory}.
\newblock springer.

\bibitem[\protect\astroncite{Yu and Xu}{2014}]{JianYu2014Categorization}
Yu, J. and Xu, Z. (2014).
\newblock Categorization axioms for clustering results.
\newblock {\em eprint arXiv:1403.2065}.

\end{thebibliography}

\appendix

\end{document}